\documentclass[10pt,twocolumn,letterpaper]{article}

\usepackage[pagenumbers]{iccv} 

\definecolor{iccvblue}{rgb}{0.21,0.49,0.74}
\usepackage[pagebackref,breaklinks,colorlinks,allcolors=iccvblue]{hyperref}

\usepackage{multirow}
\usepackage{algorithm}
\usepackage{algorithmicx}
\usepackage{algpseudocode}
\usepackage{cuted}
\usepackage{array}
\usepackage{bbding}

\def\paperID{**} 
\def\confName{**}
\def\confYear{**}

\title{Restoring Forgotten Knowledge in Non-Exemplar Class Incremental Learning through Test-Time Semantic Evolution}

\author{Haori Lu$^1$, Xusheng Cao$^1$, Linlan Huang$^1$, Enguang Wang$^1$, Fei Yang$^{2, 1}$, Xialei Liu$^{2, 1}$\\
$^1$VCIP, CS, Nankai University \qquad $^2$NKIARI, Shenzhen Futian\\
{\tt\small \{luhaori, caoxusheng, huanglinlan, enguangwang\}@mail.nankai.edu.cn, }
\and
{\tt\small \{feiyang, xialei\}@nankai.edu.cn}
}

\begin{document}
\maketitle
\begin{abstract}
Continual learning aims to accumulate knowledge over a data stream while mitigating catastrophic forgetting.
In Non-exemplar Class Incremental Learning (NECIL), forgetting arises during incremental optimization because old classes are inaccessible, hindering the retention of prior knowledge.
To solve this, previous methods struggle in achieving the stability-plasticity balance in the training stages.
However, we note that the testing stage is rarely considered among them, but is promising to be a solution to forgetting.
Therefore, we propose RoSE, which is a simple yet effective method that \textbf{R}est\textbf{o}res forgotten knowledge through test-time \textbf{S}emantic \textbf{E}volution. 
Specifically designed for minimizing forgetting, RoSE is a test-time semantic drift compensation framework that enables more accurate drift estimation in a self-supervised manner.
Moreover, to avoid incomplete optimization during online testing, we derive an analytical solution as an alternative to gradient descent.
We evaluate RoSE on CIFAR-100, TinyImageNet, and ImageNet100 datasets, under both cold-start and warm-start settings.
Our method consistently outperforms most state-of-the-art (SOTA) methods across various scenarios, validating the potential and feasibility of test-time evolution in NECIL.
\end{abstract}    
\section{Introduction}
\label{sec:intro}

Current deep learning methods~\cite{dosovitskiy2021an,he2016deep,liu2022convnet} generally assume static training data, requiring the model to be retrained on all data when data updates.
Ideally, models should adapt to new data without revisit all past data. 
But training exclusively on new data often leads to rapid degradation of prior knowledge, a phenomenon known as catastrophic forgetting~\cite{mccloskey1989catastrophic}.
Continual learning (CL) investigates how models can learn from a sequential task stream while mitigating catastrophic forgetting.
This paper focuses on class-incremental learning (CIL), where the tasks to be learned have distinct classes.
A subset of CIL methods, known as replay-based methods~\cite{bang2021rainbow,hou2019learning,rebuffi2017icarl}, store exemplars from previous tasks and include them during training on new tasks to reduce forgetting.
However, due to privacy concerns, storing exemplars is often impractical.
In more realistic scenarios, only an old checkpoint can be accessible.
Thus, we focus on class-incremental learning without exemplar storage, referred to as Non-exemplar Class Incremental Learning (NECIL)~\cite{zhu2022self, petit2023fetril, li2024fcs}.

\begin{figure}
    \centering
    \includegraphics[width=0.95\linewidth]{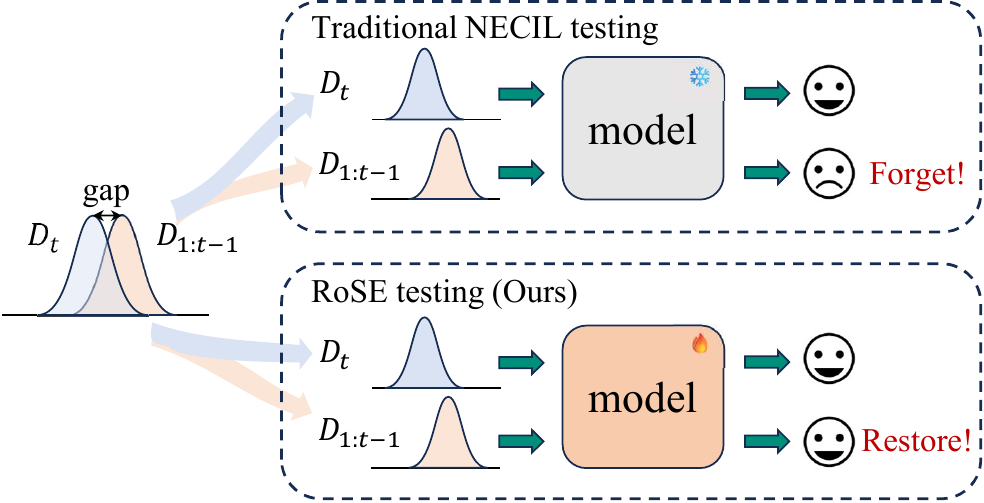}
    \caption{
    In the training stages of NECIL, the model only learns from the new task data distribution, which has a gap with the previous data distribution.
    Traditional NECIL testing cannot update the model with the test distribution to avoid forgetting the old data distribution.
    We propose to utilize test data during each incremental task's testing to restore knowledge forgotten in the training stages.
    }
    \label{teaser}
\end{figure}

In NECIL, forgetting arises because data from previous tasks is unavailable, as illustrated in \cref{teaser}.
For this issue, previous methods limit network updating in training stages to preserve stability-plasticity balance by regularization~\cite{li2017learning, magistrielastic} or even freezing the backbone~\cite{goswami2024fecam, petit2023fetril}.
However, we note that the testing stages are often overlooked by them, but have strong potential to ease the stability-plasticity dilemma.
An intuition is that, during each testing stage, test data inherently include features from previous classes, tackling the main challenge in NECIL: \textit{the inaccessibility of old data distribution}.
In this case, the model can both retain old knowledge through stability and restore it through active learning test data.
Therefore, in this paper we explore a test-time path to overcome forgetting.

To train the model at test-time, we need to design an appropriate optimization objective for test data first, which is referred to as the auxiliary task~\cite{liu2021ttt++}.
During testing, we optimize only the auxiliary task, which must satisfy two key conditions:
\textbf{1) Self-supervised.} Since test data have no labels, the auxiliary task should be effective without labels; \textbf{2) Online.} Due to test-time requirements, repeated visits of test samples are not allowed.
Therefore, the training must be performed in an online manner.

For the first condition, reducing forgetting with unlabeled data is challenging.
Since forgetting affects the prediction of old classes, the pseudo-labels are of limited quality. Incorrect pseudo-labels further lead to error accumulation. To address this, we identify an appropriate self-supervised target that avoids relying on pseudo-labels: semantic drift compensation.
As the network is trained, features from old classes drift in feature space, while the classifier fails to account for this drift, leading to wrong prediction.
Test data, which include old features, enable a comprehensive estimation of semantic drift. 
Thus, we revisit semantic drift compensation from the test-time perspective. Inspired by~\cite{li2024fcs,gomez2025exemplar}, we set our auxiliary task as training a projector to capture this drift and propose our test-time semantic evolution framework.

For the second condition, we find that the optimal solution to our auxiliary task can be derived analytically in a single step. This approach eliminates the risk of incomplete optimization typically associated with gradient descent in online training.
When test data are limited at the beginning of the test, we generate enough Gaussian pseudo-features to ensure that feature matrices are full-rank and invertible for the analytical solution.

Extensive experiments show that incorporating test-time semantic evolution into a simple baseline (LwF~\cite{li2017learning} + supervised contrastive learning~\cite{khosla2020supervised}), outperforms most state-of-the-art NECIL methods in both cold-start and warm-start settings, highlighting the potential of investing test-time mechanism in continual learning.
We hope our work could inspire the community to explore test data as a means of reducing forgetting.
In human lifelong learning, everything we encounter contributes to our memory in real time.
Similarly, continual learning models should transcend traditional constraints and leverage any available input to constantly reinforce their knowledge.
Our contributions can be summarized as:
\begin{itemize}
    \item We highlight that the testing stages can help reduce forgetting. We propose a novel test-time semantic evolution method, RoSE, which is based on a tailored auxiliary task designed to restore forgotten knowledge.
    
    \item During test-time, we compute the analytical solution rather than gradient for updating the model, addressing the limitations of online training.

    \item Extensive experiments show that RoSE outperforms most state-of-the-art NECIL methods across various scenarios. In the cold-start 10 tasks setting, RoSE surpasses SOTA methods by 7\%, 4.94\%, and 12.3\% in CIFAR100, TinyImageNet, and ImageNet100, respectively.
\end{itemize}

\section{Related work}
\label{sec:related_work}

\subsection{Class incremental learning (CIL)}
Existing class incremental learning methods can be categorized into three groups: 
\textit{Replay-based} methods~\cite{bang2021rainbow,hou2019learning,rebuffi2017icarl,shin2017continual,zhai2021hyper} save exemplars and add them to the new task training to overcome forgetting. 
\textit{Regularization-based} methods~\cite{aljundi2018memory,douillard2020podnet,kirkpatrick2017overcoming, li2017learning,liu2018rotate,wu2019large,zenke2017continual,xie2024early} constrain model updating by some regularization terms such as knowledge distillation~\cite{douillard2020podnet,li2017learning, wu2019large} or parameter importance~\cite{aljundi2018memory,kirkpatrick2017overcoming,liu2018rotate,zenke2017continual}. 
\textit{Architecture-based} methods~\cite{fernando2017pathnet,mallya2018packnet,mallya2018piggyback,serra2018overcoming,yan2021dynamically} assign a part of the network capacity to each task by modifying network layers~\cite{fernando2017pathnet,yan2021dynamically} or parameters~\cite{mallya2018packnet,serra2018overcoming,mallya2018piggyback, zhu2022self}.

In NECIL, storing exemplars to mitigate forgetting is infeasible.
Alternative methods like PASS~\cite{zhu2021prototype} use prototype replay instead of real data, while IL2A~\cite{zhu2021class} introduces dual data augmentation. FeTrIL~\cite{petit2023fetril} and FeCAM~\cite{goswami2024fecam} address forgetting by freezing the backbone after the initial task.
Some recent continual learning methods~\cite{marouf2023rethinking,singh2024controlling} investigate utilizing test data, but still follow adaptation-based methods to adapt the pre-train model.
TTACIL~\cite{marouf2023rethinking} optimizes normalization layers via entropy minimization, similar to TENT~\cite{wang2020tent}, while DoSAPP~\cite{singh2024controlling} uses teacher model pseudo-labeling to enhance generalization to old knowledge.
Instead, we propose a more appropriate test-time auxiliary task for CIL: test-time semantic drift compensation.

\subsection{Drift compensation in CIL}
Some works~\cite{yu2020semantic,goswami2024resurrecting,li2024fcs,gomez2025exemplar} mitigate forgetting by compensating for semantic drift that occurs as the model undergoes continual updates.
SDC~\cite{yu2020semantic} estimates the drift of old prototypes using weighted averaging of the feature drift. ADC~\cite{goswami2024resurrecting} adversarially perturbs current samples to old prototypes to obtain drift.
FCS~\cite{li2024fcs} and LDC~\cite{gomez2025exemplar} predict drift by training a linear projection from the old feature space to the new feature space. 
PPE~\cite{li2024progressive} evolves prototype per class directly using features per class, which requires image labels for matching prototypes. 
These methods rely on training data, but we argue that test data provide a fuller view of semantic drift. 
Thus, we select self-supervised drift compensation as a test-time auxiliary task.
Based on this, we design an online test-time semantic evolution pipeline that reduces forgetting by more effectively estimating the drift.

\section{Method}\label{sec:method}
In \cref{sec:baseline}, we outline the Non-exemplar Class Incremental Learning setting and present a simple continual learning baseline with two classical techniques to mitigate forgetting during training stages.
In \cref{sec:aux}, we revisit semantic drift compensation from the test-time perspective to design an auxiliary task aimed at reducing forgetting and propose the test-time semantic evolution method.
In \cref{sec:analytic}, we address the limitations of online optimization during test-time by an analytical solution. Finally, we present the overall algorithm in \cref{alg:overall}.

\subsection{Preliminary}\label{sec:baseline}
\paragraph{Non-Exemplar Class Incremental Learning (NECIL).} In NECIL, the overall dataset $D$ is divided into $T$ tasks, denoted by $D=\{D_t\}^T_{t=1}$. Each task dataset $D_t$ contains a set of classes $C_t$, with no overlap between classes of any two tasks: $C_i \cap C_j = \emptyset, \forall i, j \in 1\colon T$. 
During training, we update the feature extractor $f$ (parameterized by $\theta$) to extract features from images, and the classification head $h$ (parameterized by $\phi$) to predict labels.
The objective is to achieve accurate classification across all seen tasks, approximately formulated as:
\begin{equation}
    \mathop{argmin}_{\theta, \phi} \mathop{\sum}_{t=1}^T \mathbb{E}_{(x, y)\sim D_t}\mathcal{L}_{ce}(h(f(x)), y),
\end{equation}
where $(x, y)$ represents the images and labels in $D_t$, and  $\mathcal{L}_{ce}$ denotes the cross-entropy loss.
To reduce classifier forgetting, several methods~\cite{goswami2024fecam, goswami2024resurrecting, gomez2025exemplar} employ the Nearest Class Mean (NCM) classification instead of a standard classifier for predictions.
NCM computes a prototype $p_c$ for each class $c\in C_t$ after training on task $t$ by:
\begin{equation}\label{eqn:cal_proto}
    p_c = \frac{1}{N_c} \mathop{\sum}_{n=1}^{N_c} f(x_n^c), \forall c \in C_t,
\end{equation}
where $N_c$ represents the number of images in class $c$, and $x_n^c$ is the $n$-th image in class $c$.
The prediction is made by identifying the class whose prototype has the minimum cosine distance from the image feature.
Our method also adopts NCM to predict:
\begin{equation}
    \hat{y} = \mathop{argmin}_c \text{cos}\langle f(x), p_c\rangle.
\end{equation}

\paragraph{A simple NECIL baseline.}
Our baseline incorporates knowledge distillation from LwF~\cite{li2017learning}, which aligns the predicted probabilities of old classes in the current model with those of the previous model. The loss function is:
\begin{equation}
    \mathcal{L}_{KD} = -\sum_{i=1}^{|C_{1:t-1}|} \hat{y}_{t-1}^i*log(\hat{y}_t^i),
\end{equation}
where $\hat{y}_{t-1}^i$ and $\hat{y}_{t}^i$ denote the predicted probabilities for class $i$ from the models $f_{t-1}$ and $f_t$.
To learn a more generalized feature space, we apply the supervised contrastive loss~\cite{khosla2020supervised} during training. This loss term is defined as:
\begin{equation}
    \mathcal{L}_{SCL}=-\mathop{\sum}_{i=1}^N\mathop{\sum}_{k^+\in P(x_i)} log\frac{exp(z_i^\top k^+/\tau)}{\mathop{\sum}_{k'\in N(x_i)}exp(z_i^\top k'/\tau)},
\end{equation}
where $x_i$ is the $i$-th image, $z_i=f(x_i)$ is the feature extracted from $x_i$, $P(x_i)$ and $N(x_i)$ represent the positive and negative set for $x_i$. 
$P(x_i)$ includes features from the same class as $x_i$ within this batch, while $N(x_i)$ comprises features from other classes in the batch.
These losses are combined with the cross-entropy loss $\mathcal{L}_{ce}$ to form the complete loss function for the baseline:
\begin{equation}
    \mathcal{L}_{base} = \mathcal{L}_{ce} + \lambda_1 \mathcal{L}_{KD} + \lambda_2 \mathcal{L}_{SCL}.
\end{equation}

\begin{figure*}
    \centering
    \includegraphics[width=0.9\linewidth]{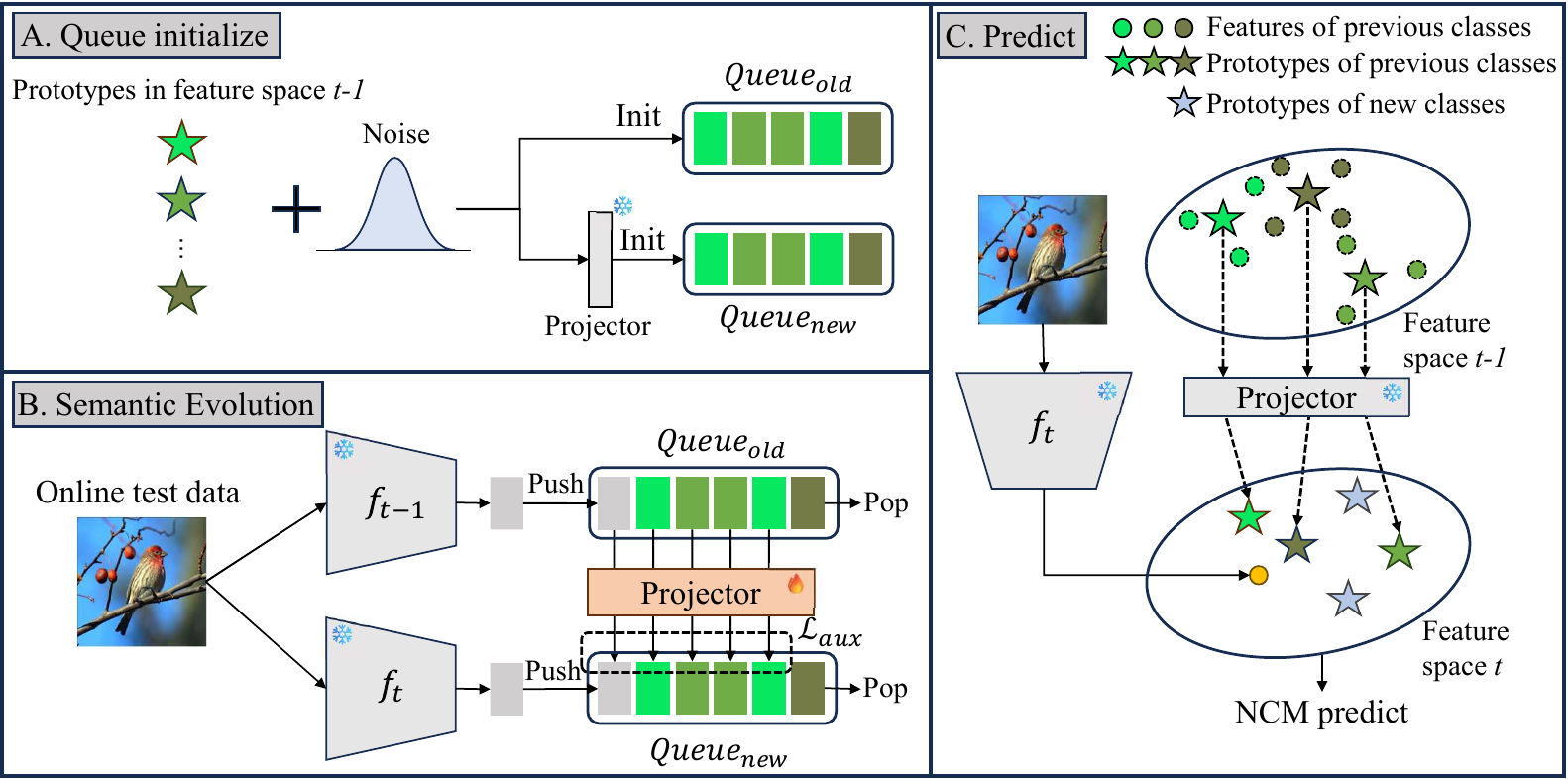}
    \caption{Test process of RoSE. We maintain two feature queues, $Q_{old}$ and $Q_{new}$ during testing.
    \textbf{A)} At test start, we generate sufficient pseudo-features to initialize $Q_{old}$. These pseudo-features are then fed into the projector to initialize $Q_{new}$.
    \textbf{B)} During each incremental task's testing, the projector is updated by online test data to better evolve the features and prototypes.
    \textbf{C)} Through the updated projector, we estimate the positions of old prototypes in the new feature space for image classification.
    }
    \label{fig:method}
\end{figure*}

\subsection{Test-time semantic evolution}\label{sec:aux}
During the training phase, our method adheres to the aforementioned baseline, and in the testing stage, we propose a novel test-time semantic evolution strategy to replace the conventional test.
To facilitate clarify, we follow \cite{liu2021ttt++} to decompose the optimization process into two components: the main task and the auxiliary task.
The main task corresponds to the primary target, which can be considered as optimizing $\mathcal{L}_{base}$ in NECIL.
The auxiliary task is a self-supervised task trained alongside the main task during training while updating the network independently during testing. 
In NECIL, we expect the testing stages could reduce forgetting.
Thus, we explore a new auxiliary task specifically designed to reduce forgetting in NECIL.

\paragraph{Revisiting semantic drift compensation from the test-time perspective.}
As the model evolves incrementally within the data stream, features of historical data extracted by the updated model will drift compared to those extracted earlier.
Crucially, the inaccessibility of old data prevents the classifier (in our method it's the prototype of old classes) from adapting to this representational shift.
This misalignment is a significant source of forgetting. 
While some methods~\cite{li2024fcs,gomez2025exemplar} capture drift by training a projector $P$ to project the old feature space into the new feature space, their training regime remains suboptimal. 
Let $W$ denotes the parameters of $P$.
The optimization objective for $W$ is:

\begin{equation}\label{eqn:projector}
    W^* = \mathop{argmin}_{W}\mathbb{E}_{z^{t-1} \sim f_{t-1}(D_t)\atop z^{t} \sim f_{t}(D_t)}||P(z^{t-1}) - z^t ||_2.
\end{equation}

\cref{eqn:projector} reveals a critical limitation in prior approaches: The projector is exclusively trained on the current task’s training data $D_t$, failing to leverage the wealth of historical features embedded in test data.
Therefore, we revisit semantic drift compensation from the test-time perspective.
During testing, the model encounters test data containing previously seen classes.
We can concurrently extract paired features $\{z_{test}^{t-1}, z_{test}^t\}$ by forwarding them into both $f_t$ and $f_{t-1}$.
This test information enables the projector to capture more precise drifts of old classes.
Since a prototype represents a class centroid in the embedding space, its drift directly correlates with the class-conditional feature drift.
This makes semantic drift compensation an ideal auxiliary task: it is a self-supervised task that mitigates forgetting using only test features extracted from $f_{t-1}$ and $f_t$.
The test-time semantic evolution process is illustrated in \cref{fig:method}B.
The auxiliary loss function is:

\begin{equation}\label{eqn:projector_test}
    \mathcal{L}_{aux} = \mathbb{E}_{z^{t-1}_{test} \sim f_{t-1}(D_t^{test})\atop z^{t}_{test} \sim f_{t}(D_t^{test})}||P(z^{t-1}_{test}) - z^{t}_{test}) ||_2.
\end{equation}

An additional benefit of our method is that the model should naturally be most oriented towards higher accuracy for test data. By evolving semantics at test-time, the model dynamically prioritize most attention to performance on the current test data.
Thus, for more common data, the model more accurately estimates the drift of relevant prototypes, while for less common data, it allocates less attention.
This leads to an intelligent “active forgetting”~\cite{wang2023incorporating} phenomenon, where the model selectively forgets less frequent knowledge to free up capacity for more common knowledge.
For prediction, the projector $P$ projects all old prototypes $p^{t-1}$ onto $p^t$, effectively evolving the prototypes (\cref{fig:method}C).
The projection process is computed as:
\begin{equation}\label{eqn:updateP}
    p_c^{t} = P\left(p_c^{t-1} \right), \forall c \in C_{1:t-1}.
\end{equation}

\begin{table*}[]
\centering
\begin{tabular}{lcccccccc}
\toprule
 \multirow{2}{*}{Methods} & \multicolumn{3}{c}{CIFAR100} & \multicolumn{3}{c}{TinyImageNet} & ImageNet100 \\
 & \multicolumn{1}{c}{5 tasks} & \multicolumn{1}{c}{10 tasks} & \multicolumn{1}{c}{20 tasks} & \multicolumn{1}{c}{5 tasks} & \multicolumn{1}{c}{10 tasks} & \multicolumn{1}{c}{20 tasks} & \multicolumn{1}{c}{10 tasks} \\ \midrule
LwF~\cite{li2017learning} & 45.35 & 26.14 & 16.54 & 38.81 & 27.42 & 11.87 & 29.12 \\
SDC~\cite{yu2020semantic} & 54.94 & 41.36 & 33.46 & 40.05 & 27.15 & 11.03 & 32.90 \\
PASS~\cite{zhu2021prototype} & 49.75 & 37.78 & 26.32 & 36.44 & 26.58 & 15.97 & 39.16 \\
FeTrIL~\cite{petit2023fetril} & 45.11 & 36.69 & 26.90 & 29.91 & 23.88 & 16.29 & 40.26 \\
FeCAM$_{1\,cov}$~\cite{goswami2024fecam} & 44.15 & 31.81 & 24.26 & 34.09 & 25.98 & 18.77 & 39.94 \\
FeCAM$_{c\,cov}$~\cite{goswami2024fecam} & 47.28 & 33.82 & 26.08 & 25.62 & 23.21 & 22.24 & 43.00 \\
EFC~\cite{magistrielastic} & 53.07 &43.62 & 32.15 & 40.29 & 34.10  & \underline{28.69} &-  \\
ADC~\cite{goswami2024resurrecting} & \underline{59.14} & \underline{46.48} & 35.49 & 41.02 &32.32  & 21.33 & 47.58 \\
LDC~\cite{gomez2025exemplar} & 58.33 & 45.52 & \underline{36.91} & \underline{44.35} & \underline{34.99} & 24.95 & \underline{50.88}  \\ \midrule
 RoSE (Ours) & \textbf{63.10} &\textbf{53.48}  & \textbf{44.10} & \textbf{47.07} & \textbf{39.93} & \textbf{31.57} & \textbf{63.18} \\
\bottomrule
\end{tabular}
\caption{
Comparison with other methods on cold-start settings.
\textbf{Bold} indicates the best result. \underline{Underline} indicates the second-best result.
}
\vspace{-0.2cm}
\label{tab:b0}
\end{table*}

\subsection{Analytical optimization for online learning}\label{sec:analytic}

Another critical consideration is that \cref{eqn:projector_test} must be implemented online due to the constraints of test-time. 
Employing gradient descent for this process may lead to incomplete optimization, particularly when the quantity of test data is limited.
RoSE circumvents this issue effectively because, in our testing, only the projector $P$ is trainable, which consists of only one linear layer. This simplicity enables computing the analytical solution $W^*$ for $P$ directly.
Specifically, during testing, we maintain two queues, $Q_{old}$ and $Q_{new}$.
For each test image $x_i$, we extract the features $z_i^{t-1}$ and $z_i^t$ from $f_{t-1}$ and $f_t$, respectively, and store these features in $Q_{old}$ and $Q_{new}$.
\cref{eqn:queue} illustrates $Q_{old}$ and $Q_{new}$ in matrix form, with $n$ rows and $d$ columns. $n$ denotes the queue capacity and $d$ represents the feature dimension.
\begin{equation}\label{eqn:queue}
    Q_{old} = 
    \begin{pmatrix}
    z^{t-1}_{11}  & ... & z^{t-1}_{1d} \\
    \vdots & \ddots &\vdots \\
    z^{t-1}_{n1}  & ... & z^{t-1}_{nd} 
    \end{pmatrix}\!, 
    Q_{new} = 
    \begin{pmatrix}
    z^{t}_{11}  & ... & z^{t}_{1d} \\
    \vdots & \ddots &\vdots \\
    z^{t}_{n1}  & ... & z^{t}_{nd} 
    \end{pmatrix}\!.
\end{equation}
We expect the projection can fulfill:
\begin{equation}
    Q_{old}  W = Q_{new}.
\end{equation}
Hence, the analytical solution $W^*$ for $W$ is:
\begin{equation} \label{eqn:analyticalW}
    W^* = \left(Q_{old}^\top\thinspace Q_{old} \right)^{-1} Q_{old}^\top\thinspace Q_{new}.
\end{equation}

Since the loss function in \cref{eqn:projector_test} is convex, this analytical solution ensures optimality.
However, in the early testing, the number of test samples is inadequate, leading to a rank-deficient and non-invertible $(Q_{old}^\top\thinspace Q_{old})$.
Therefore, at test start, we generate-pseudo features $z^{t-1}_{pseudo}$ by adding Gaussian noise to old prototypes $p_{t-1}$, which are then used to initialize $Q_{old}$, 
and compute $z^t_{pesudo}=P(z^{t-1}_{pseudo})$ to initialize $Q_{new}$ (\cref{fig:method}A).
The formula for generating a $z^{t-1}_{pesudo}$ is
 \cref{eqn:generate}, where $c$ is a randomly chosen old class.

\begin{equation}\label{eqn:generate}
   z^{t-1}_{pesudo} = p_c^{t-1} + \alpha \cdot \mathcal{N}\sim \left(0, 1\right).
\end{equation}

As the test goes on, the queue continues to update in an online manner.
Whenever new test data arrives, we append the new features to the end of the queue and pop the same number of features from the start of the queue.
We outline the overall algorithm for RoSE in \cref{alg:overall}.

\vspace{-0.2cm}
\begin{algorithm}[!h]
    \caption{Overall Algorithm during task $t$}
    \label{alg:overall}
    \renewcommand{\algorithmicrequire}{\textbf{Input:}}
    \renewcommand{\algorithmicensure}{\textbf{Output:}}
    \renewcommand{\algorithmiccomment}[1]{\hfill $\triangleright$ #1}
    \begin{algorithmic}[1]
        \Require train dataset $D_t$, test dataset $D_t^{test}$, encoder parameter $\theta_{t-1}$, projector $P$, old prototypes $p_{C_{1:t-1}}^{t-1}$, main loss $\mathcal{L}_{base}$, auxiliary loss $\mathcal{L}_{aux}$, queue budget S, total epoch $\mathcal{E}$, learning rate $\sigma$, noise coefficient $\alpha$. 
        \Ensure classification results $res$    
        
        \State  $\text{copy}: \theta_t \leftarrow \theta_{t-1}$
        \State $ \theta_t, P \leftarrow \text{TrainModel}(D_t, \theta_{t}, \theta_{t-1}, P, \mathcal{L}_{base}, \mathcal{L}_{aux}, \mathcal{E}, \sigma)$ \Comment{Training stages}
        \State $p_C^t \leftarrow \text{ComputeNewPrototypes}(D_t, \theta_{t} ) $\Comment{\cref{eqn:cal_proto}}
        \State $Q_{old}=[~], Q_{new}=[~], res=[~]$ \Comment{Test start}
        \State $Q_{old}, Q_{new} \leftarrow \text{InitQueue}(Q_{old}, Q_{new}, P, S, p_{C_{1:t-1}}^{t-1}, \alpha)$ \Comment{\cref{fig:method}A}
        \State $res \leftarrow \text{TestSemanticEvolution}(D_t^{test}, Q_{old}, Q_{new}, \theta_{t},$ 
        \Statex $\theta_{t-1}, \mathcal{L}_{aux}, p_{C_{1:t-1}}^{t-1}, p_C^t)$ \Comment{\cref{fig:method}B and C}
        \State \Return $res$
    \end{algorithmic}
\end{algorithm}
    \vspace{-0.3cm}
\section{Experiments}
\label{sec:experiments}

\subsection{Experimental setups}

\paragraph{Datasets.} 
We conduct experiments on three datasets: CIFAR100~\cite{krizhevsky2009learning}, 
TinyImageNet~\cite{le2015tiny},
and ImageNet100~\cite{russakovsky2015imagenet}.
For a detailed description of these datasets, please refer to \textit{Supplementary Materials (SM)}.
We evaluate both cold- and warm-start settings.
In cold-start settings, classes are equally divided across all tasks.
In warm-start settings, the first task contains half of the total classes, while the remaining tasks equally divide the rest.
For the CIFAR100 20 tasks warm-start, the first task contains 40 classes, and subsequent tasks contain 3 classes.

\begin{table*}[]
\centering
\begin{tabular}{lccccccc} \toprule
\multirow{2}{*}{Methods} & \multicolumn{3}{c}{CIFAR100} & \multicolumn{3}{c}{TinyImageNet} & ImageNet100 \\
 & 5 tasks & 10 tasks & 20 tasks & 5 tasks & 10 tasks & 20 tasks & 10 tasks \\ \midrule
LwF~\cite{li2017learning} & 24.01 & 16.52 & 14.66 & 14.73 & 7.60 & 3.11 & 13.70 \\
SDC~\cite{yu2020semantic} & 43.26 & 34.35 & 30.15 & 27.09 & 12.72 & 7.89 & 18.52 \\
PASS~\cite{zhu2021prototype} & 56.40 & 50.69 & 46.93 & 42.52 & 40.27 & 34.80 & 54.50 \\
IL2A~\cite{zhu2021class} & 53.93 & 45.76 & 44.24 & 39.53 & 36.55 & 30.02 &  - \\
SSRE~\cite{zhu2022self} & 56.97 & 56.57 & 51.92 & 41.45 & 41.18 & 41.03 & 59.32 \\
FeTrIL~\cite{petit2023fetril} & 58.12 & 57.64 & 52.48 & 42.92 & 42.41 & 41.33 & 61.22 \\
FeCAM$_{1\,cov}$~\cite{goswami2024fecam} & 59.03 & 59.03 & 55.27 & 46.50 & 46.50 & 46.50 & 67.14 \\
FeCAM$_{c\,cov}$~\cite{goswami2024fecam} & 62.10 & \underline{62.10} & \underline{58.58} & 50.48 & \underline{50.48} & \textbf{50.48} & \textbf{70.90} \\
EFC~\cite{magistrielastic} & 61.87 & 60.87 & 55.71 & \underline{51.61} & 50.40 & 48.68 & - \\
ADC~\cite{goswami2024resurrecting} & 49.49 & 44.08 & 35.60 & 38.61 & 24.39 & 21.18 & 36.96 \\
LDC~\cite{gomez2025exemplar} & 51.61 & 44.79 & 32.16 & 39.78 & 26.72 & 19.64 & 42.88 \\ 
FCS~\cite{li2024fcs} & \underline{62.13} & 60.39 & 58.36 & 46.04 & 44.95 & 42.57 & 61.76  \\ 
 \midrule
RoSE (Ours) & \textbf{66.70} & \textbf{64.46} & \textbf{59.54} & \textbf{54.64} & \textbf{51.11} & \underline{50.07} & \underline{69.62} \\
\bottomrule
\end{tabular}
\caption{
Comparison with other methods on warm-start settings.
\textbf{Bold} indicates the best result. \underline{Underline} indicates the second-best result.}
\label{tab:b50}
\end{table*}

\paragraph{Evaluation metric.}
We report the last accuracy for evaluation. Last accuracy is the average accuracy across all classes after the model has learned on all tasks. We also report accuracy after each task in \cref{fig:acc_stage} to show the intermediate task results of methods.

\paragraph{Implementation details.}
The backbone of our method is ResNet-18~\cite{he2016deep}. 
For the warm-start CIFAR100 and TinyImageNet settings, we follow the data augmentation strategies in FCS~\cite{li2024fcs}.
We use the Adam optimizer with a learning rate of 0.001 to optimize the projector.
The hyperparameter $\lambda_1$ is set to 10 for the CIFAR-100, TinyImageNet, and 5 for ImageNet100 following LDC~\cite{gomez2025exemplar}. $\lambda_2$ is set to 0.1 for 5 tasks and 10 tasks settings, and 0.03 for the 20 tasks and ImageNet experiments. 
$\alpha$ is set to 0.02.
The test dataset is shuffled. For other training details, please refer to \textit{SM}.

\subsection{Comparison with SOTA methods}

\paragraph{Comparison methods.}
We compare RoSE with several SOTA NECIL methods.
Since the SOTA methods differ between cold-start and warm-start settings, the compared methods vary accordingly.
In cold-start, we compare methods including LwF~\cite{li2017learning}, SDC~\cite{yu2020semantic}, PASS~\cite{zhu2021prototype}, FeTrIL~\cite{petit2023fetril}, FeCAM~\cite{goswami2024fecam}, EFC~\cite{magistrielastic}, ADC~\cite{goswami2024resurrecting}, LDC~\cite{gomez2025exemplar}.
In warm-start, we compare methods including LwF~\cite{li2017learning}, SDC~\cite{yu2020semantic}, PASS~\cite{zhu2021prototype}, IL2A~\cite{zhu2021class}, SSRE~\cite{zhu2022self}, FeTrIL~\cite{petit2023fetril}, FeCAM~\cite{goswami2024fecam}, EFC~\cite{magistrielastic}, FCS~\cite{li2024fcs}, ADC~\cite{goswami2024resurrecting}, LDC~\cite{gomez2025exemplar}.

\paragraph{Comparison with cold-start SOTA methods.}
\cref{tab:b0} shows the results of RoSE compared to existing SOTA methods in cold-start settings. 
RoSE shows significant improvements, dramatically outperforms existing SOTA methods with absolute gains of 3.96\%, 7\%, and 7.19\% on CIFAR100, 2.72\%, 4.94\%, and 2.88\% on TinyImageNet, and 12.3\% on ImageNet100, proving the advantages of our test-time semantic evolution in mitigating forgetting.
The poor performance of PASS, FeTrIL and FeCAM proves that the warm-start methods are not suitable for cold-start settings.
In \cref{fig:acc_stage}, we show the accuracy of SOTA methods at each step in the 10 tasks settings.
RoSE consistently outperforms others, with the performance gap essentially widening as the number of tasks increases.

\paragraph{Comparison with warm-start SOTA methods.}
\cref{tab:b50} shows a comparison of RoSE’s performance with the current SOTA methods in warm-start settings.
On CIFAR100, RoSE outperforms existing methods by 4.57\%, 2.36\%, and 0.96\%. For TinyImageNet, RoSE achieves improvements of 3.03\% and 0.63\% in the 5 tasks and 10 tasks settings, respectively.
On TinyImageNet 20 tasks and ImageNet, RoSE performs slightly below FeCAM$_{c\,cov}$~\cite{goswami2024fecam}.
This is because semantic drift is less pronounced during warm-start, as other drift compensation methods (SDC~\cite{yu2020semantic}, ADC~\cite{goswami2024resurrecting}, and LDC~\cite{gomez2025exemplar}) also show inferior superiority.
Nevertheless, RoSE's overall performance surpasses FeCAM$_{c\,cov}$, especially in cold-start settings.
This is because FeCAM$_{c\,cov}$ freezes the encoder after training the first task, thus requiring a large number of classes in the first task.
Furthermore, FeCAM$_{c\,cov}$ stores a covariance matrix per class, which increases the storage as the number of classes increases.
For example, on TinyImageNet, it stores up to 52.4M parameters for the covariance matrix (200$\times$512$\times$512).
FeCAM$_{1\,cov}$ preserves only one common covariance matrix, eliminating the need for increased storage, but it's still clearly worse than our method in all settings.
Thus, RoSE remains competitive with FeCAM.

\subsection{Ablation study}

\begin{figure}
    \centering
    \includegraphics[width=0.95\linewidth]{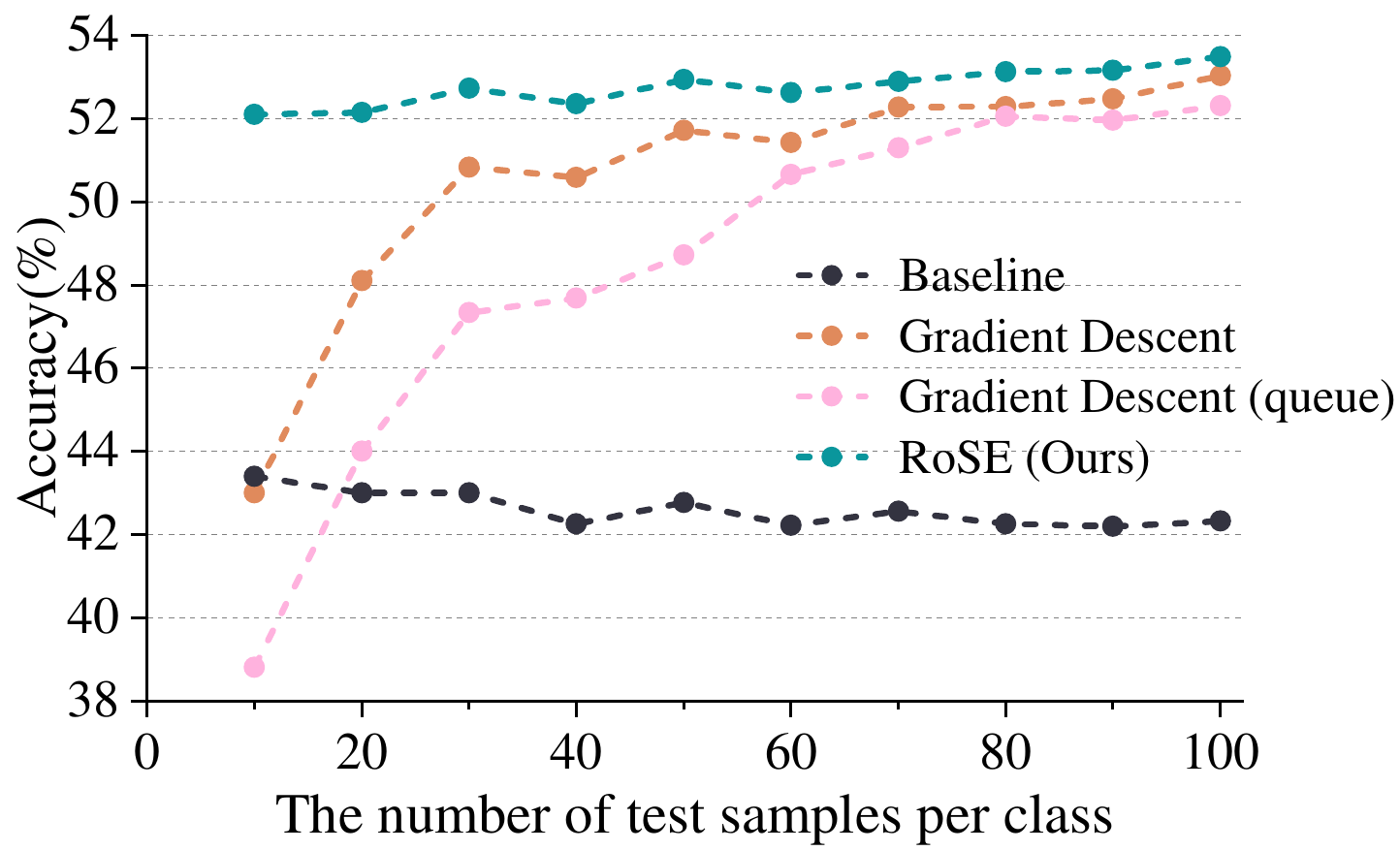}
    \caption{
    Relation between the accuracy and the number of test samples per class for baseline, gradient descent, gradient descent with a queue of the same capacity as RoSE, and RoSE.
    }
    \vspace{-0.1cm}
    \label{fig:testnum}
\end{figure}

\paragraph{Test accuracy as the number of test samples increases.}
\cref{fig:testnum} shows the accuracy as the number of test samples increases in CIFAR100 10 tasks cold-start.
The accuracy generally improves as the number of test samples increases, aligning with the intuition that more test data provides richer information.
We also include the gradient descent results to compare the analytical solution with it.
When the test number per class is small ($\textless 10$),  gradient descent performs even worse than the baseline described in \cref{sec:baseline}, indicating that insufficient iterations hinder effective model optimization.
The optimization process is interrupted before the parameters stabilize.
We also present results for gradient descent with a queue of the same capacity as RoSE, which performs worse than without the queue.
This suggests that using a queue complicates optimization, making gradient descent be even less sufficiently optimized with limited data.
As test data increases, gradient descent improves, but consistently lags behind RoSE.
In conclusion,  RoSE effectively mitigates the challenges of online test-time semantic evolution, and thus has broader applicability.

\begin{figure}
    \centering
    \includegraphics[width=0.99\linewidth]{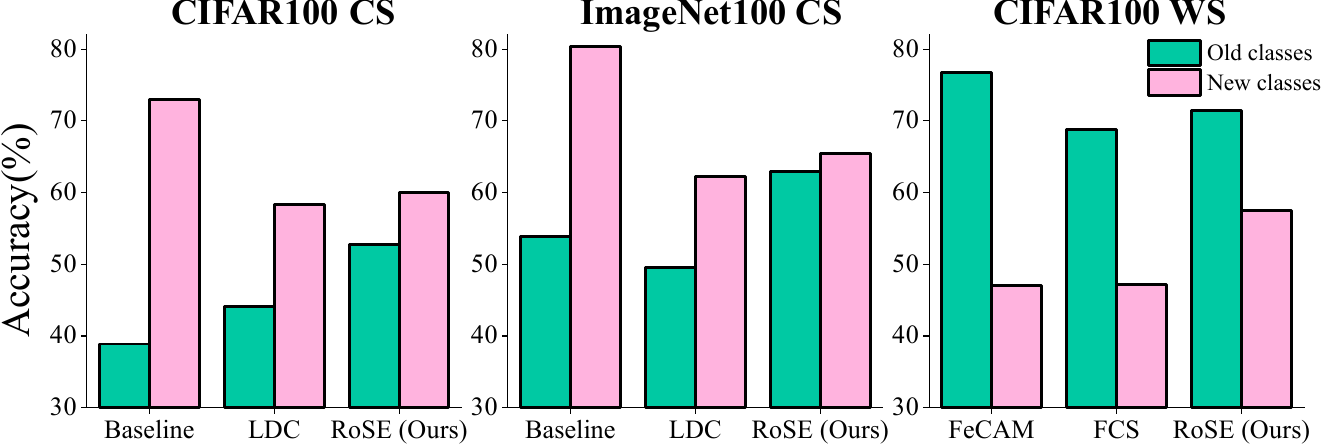}
    \caption{
    Accuracy of different methods on old and new classes. `CS', `WS' denotes `cold-start' and `warm-start'.
    }
    \label{fig:oldnew}
    \vspace{-0.2cm}
\end{figure}

\begin{figure*}
    \centering
    \includegraphics[width=0.99\textwidth]{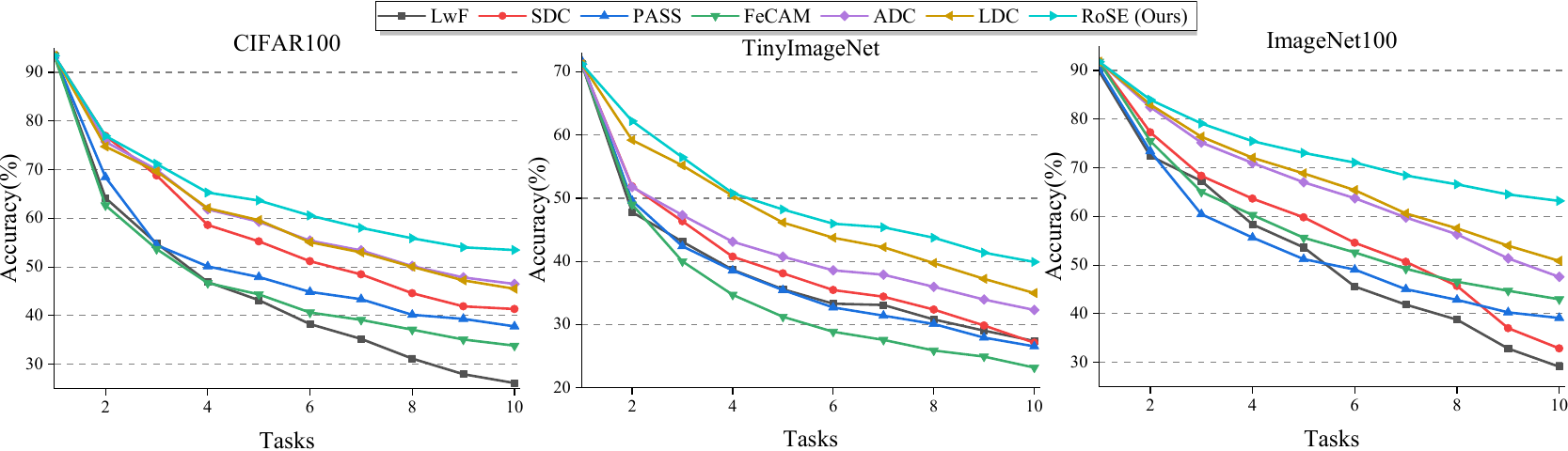}
    \caption{%
    Accuracy after each task in the cold-start 10 tasks settings.
    }
    \label{fig:acc_stage}
    \vspace{-0.2cm}
\end{figure*}

\paragraph{The accuracy of old and new classes.}
In \cref{fig:oldnew}, we show RoSE's accuracy on old and new classes for CIFAR100 and ImageNet100 under cold-start (CS) 10 tasks and CIFAR100 under warm-start (WS) 10 tasks. 
In CS, the stability of old classes is more important, so we set the `New classes' as the classes of the last task and the `Old classes' as the classes of all previous tasks.
Compared to the baseline in \cref{sec:baseline} and LDC, RoSE significantly improves old class accuracy, proving RoSE's effectiveness in restoring old task knowledge. 
In WS, the plasticity of the new classes is crucial, so we set the `Old classes' as the first task's classes and the `New classes' as all classes after the first task.
Compared to FeCAM and FCS, RoSE enhances new class knowledge retention in the final model.
Whether in CS or WS, RoSE consistently better balances the old and new classes.

\paragraph{Comparison with test-time methods.}
In certain other settings, the studied methods also update the model at test-time (see \textit{SM} for a discussion on our differences from them).
To highlight RoSE’s uniqueness, we adapt these methods to NECIL and compare them with RoSE in \cref{tab:ttt_compare}.
The baselines include Test-time adaptation (TENT~\cite{wang2020tent}), Test-time training (TTT~\cite{sun2020test}, TTT++~\cite{liu2021ttt++}), and Continual test-time adaptation (CoTTA~\cite{wang2022continual}, EATA~\cite{niu2022efficient}) methods.
Since they mainly focus on out-of-distribution adaptation, they struggle with restoring knowledge.
RoSE significantly outperforms them, underscoring the necessity of redesigning the auxiliary task specifically for overcoming forgetting. 

\begin{table}[]
\centering
\resizebox{\columnwidth}{!}{
\begin{tabular}{lccc} \toprule
\multicolumn{4}{c}{\textbf{Cold-start}} \\ \midrule
Method & CIFAR100 & TinyImageNet & ImageNet100 \\ \midrule
TENT~\cite{wang2020tent} & 13.12 & 8.22 & 31.52 \\
TTT~\cite{sun2020test} & 36.93 & 24.35 & 30.66 \\
TTT++~\cite{liu2021ttt++} & 40.16 & 31.54 & 32.32 \\
CoTTA~\cite{wang2022continual} & 28.36 & 9.65 & 14.90 \\
EATA~\cite{niu2022efficient} & 46.87 & 35.34 & 54.90 \\ \midrule
RoSE (Ours) &\textbf{ 53.48} & \textbf{39.93} & \textbf{63.18} \\ \midrule
\multicolumn{4}{c}{\textbf{Warm-start}} \\ \midrule
Method & CIFAR100 & TinyImageNet & ImageNet100 \\ \midrule
TENT~\cite{wang2020tent} & 6.07 & 7.38 & 7.62 \\
TTT~\cite{sun2020test} & 45.55 & 34.22 & 34.39 \\
TTT++~\cite{liu2021ttt++} & 54.72 & 43.32 & 61.80 \\
CoTTA~\cite{wang2022continual} & 43.92 & 40.93 & 53.40 \\
EATA~\cite{niu2022efficient} & 49.65 & 34.05 & 54.42 \\ \midrule
RoSE (Ours) & \textbf{64.46} & \textbf{51.11} & \textbf{69.62} \\ \bottomrule
\end{tabular}
}
\caption{Comparison with test-time adaptation methods.}
\label{tab:ttt_compare}
\vspace{-0.2cm}
\end{table}

\paragraph{The adaptive ability of RoSE.}
Previous experiments all use a balanced test set.
However, in real-world applications, test data is often unbalanced. 
To simulate this, we conduct unbalanced test data experiments on CIFAR100 10 tasks cold-start.
Specifically, we randomly select 50 classes, denoted as $C_{select}$, to form the unbalanced test set $D^{test}_{C_{select}}$, while the remaining 50 classes data are represented by $D^{test}_{C-C_{select}}$.
In the unbalanced setting, only $D^{test}_{C_{select}}$ is used for testing. 
From the results in \cref{tab:unbalanced}, we observe an interesting phenomenon: compared to the balanced setting, the accuracy of $D^{test}_{C_{select}}$ improves (49.08$\rightarrow$50.48) under the unbalanced distribution, while the accuracy of $D^{test}_{C-C_{select}}$ decreases (56.80$\rightarrow$54.42).
This suggests that RoSE adapts to the unbalanced distribution, focusing on reducing forgetting for more frequent features while actively sacrificing knowledge from less common classes.
\begin{table}[]
\centering
\begin{tabular}{lcc} \toprule
 & Balanced & Unbalanced \\ \midrule
Accuracy on $D^{test}_{C_{select}}$ & 49.08 & 50.48 \\
Accuracy on $D^{test}_{C-C_{select}}$ & 56.80 & (54.42) \\ \bottomrule
\end{tabular}
\caption{
Accuracy of RoSE on balanced and unbalanced test sets. ``()'' indicates that this part of the data was not actually used for test-time semantic evolution.
}
\vspace{-0.3cm}
\label{tab:unbalanced}
\end{table}

\begin{figure}
    \centering
    \includegraphics[width=0.8\linewidth]{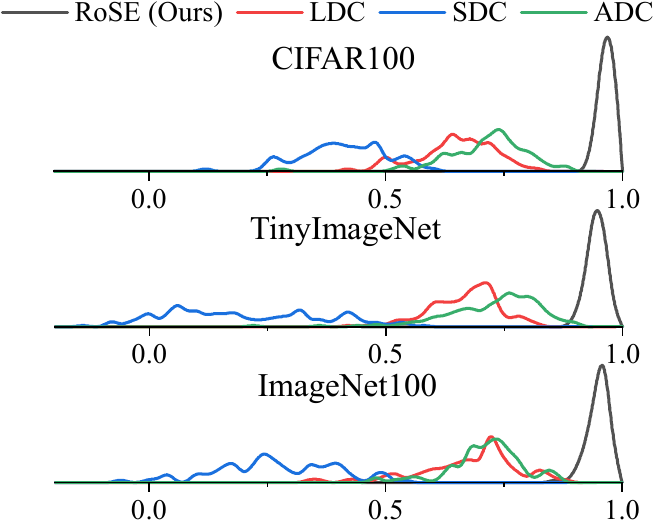}
    \caption{The distribution of cosine similarity between estimated and real drift across different methods on  CIFAR100, TinyImageNet, and ImageNet100.}
    \label{fig:driftcifar}
    \vspace{-0.2cm}
\end{figure}

\paragraph{Drift similarity between real drift and drift estimated with train data or test data.}
In \cref{fig:driftcifar}, we plot the distribution of cosine similarities between the prototype drift estimated by different methods and the real drift.
Methods we compared include SDC~\cite{yu2020semantic}, ADC~\cite{goswami2024resurrecting}, and LDC~\cite{gomez2025exemplar}.
The results are measured after the last task in the cold-start 10 tasks settings.
RoSE has significant superiority in drift similarity. On CIFAR100, almost all drift similarities of RoSE exceed 0.95, 
while the similarities of other methods are distributed around 0.35 (SDC) and 0.7 (ADC, LDC).
This indicates that RoSE can accurately estimate semantic drift and basically solves the forgetting caused by misalignment between prototypes and the encoder.

\paragraph{Accuracy as queue capacity varies.}
We conduct experiments on the CIFAR100 10 tasks cold-start setting, varying the capacity of the feature queues, $Q_{old}$ and $Q_{new}$, as shown in \cref{tab:queue_capacity}.
As the queue capacity increases, the accuracy of RoSE first improves, then starts to decline.
When the queue capacity is too small, insufficient information is available to capture semantic drift. 
However, when the queue capacity is too large, the drift estimation tends to focus on optimizing across the entire feature space, which may not be optimal for drift compensation specific to current test samples, thus the accuracy also declines.
At a moderate queue capacity, there is enough information to accurately capture drift while ensuring that the current test features' drift can also contribute enough to estimation.
Based on these observations, we set the queue capacity to 3000.

\begin{table}[]
\centering
\begin{tabular}{lccccc} \toprule
Capacity  & 500 & 1000 & 2000 & 3000 & 5000  \\ \midrule
Accuracy  & 51.45 & 53.10 & 53.37 & \textbf{53.48} & 53.11 \\ \bottomrule
\end{tabular}
\caption{
Accuracy of RoSE as the queue capacity varies.
}
\label{tab:queue_capacity}
\end{table}

\paragraph{Additional computational costs of RoSE.}
RoSE introduces additional computational costs by training a projector during testing. 
To evaluate this, we measured the average inference time per sample for the inference baseline (no training at test-time), gradient descent, and the analytical solution. 
The results in \cref{tab:inference_time} indicate that the time consumption increases successively from the baseline to gradient descent and then to the analytical solution.
This reflects the trade-off between inference time and performance.

\begin{table}[]
\centering
\begin{tabular}{lc}\toprule
 & Avg. inference time (s~/~img) \\ \midrule
Inference baseline & 0.0076 \\
Gradient descent & 0.0118 \\
Analytical solution & 0.0136 \\ \bottomrule
\end{tabular}
\caption{Comparison of average inference time.}
\label{tab:inference_time}
\vspace{-0.2cm}
\end{table}
\section{Conclusion}\label{sec:conclusion}
We notice that the testing stages are an unexplored but valuable asset for the current NECIL.
Therefore, we analyze two conditions for updating the model at test-time: self-supervise and online, and one purpose: overcoming forgetting, in NECIL.
Based on this, we designed a test pipeline for semantic evolution, which enables precise drift estimation for old prototypes when testing.
We also replace the test-time gradient descent with an analytical solution, eliminating the risk of online optimization.
Extensive experiments demonstrate that incorporating our test-time semantic evolution into a simple baseline can achieve state-of-the-art performance in both cold-start and warm-start settings, indicating that testing stages effectively restore forgotten knowledge.
Our research showcases the strong potential of testing stages for improving NECIL, and paves the way for a novel avenue of continual learning exploration.

{
    \small
    \bibliographystyle{ieeenat_fullname}
    \bibliography{main}
}
\clearpage
\setcounter{page}{1}
\setcounter{section}{0}
\maketitlesupplementary

In this supplementary material, we provide additional details as follows: a detailed description of datasets in \cref{sec:sm_datasets}, more implementation details in \cref{sec:sm_implement}, and a detailed pseudo-code in \cref{sec:code}. Regarding the potential data leakage concern, as the test data for each class remains unchanged during each test stage, we address this issue by conducting experiments in \cref{sec:sm_notsame}, demonstrating that RoSE remains effective under such conditions. We also provide more comparisons with other test-time updating methods in terms of settings (\cref{sec:relative_setting}) and experimental results (\cref{sec:more_ttt_compare}). In \cref{sec:upper}, we compare RoSE with the upper bound of using gradient descent. And we provide the average performance of RoSE over three runs in \cref{sec:sm_multiple_runs}.

\section{The detailed description of datasets}\label{sec:sm_datasets}
We conduct experiments on the following datasets: 1) CIFAR100~\cite{krizhevsky2009learning}, 
which contains 100 classes with 500 training images and 100 test images per class. The image resolution is 32$\times$32.
2) TinyImageNet~\cite{le2015tiny},
which contains 200 classes with 500 training images and 50 test images per class. The image resolution is 64$\times$64.
3) ImageNet100~\cite{russakovsky2015imagenet}, which consists of 100 classes selected from ImageNet. Each class has 1300 training images and 50 test images. The image resolution is 224$\times$224.

\section{Other implementation details}\label{sec:sm_implement}
In the cold-start settings, we use a batch size of 128 and the SGD optimizer to train the backbone. For the first task, we set the learning rate to 0.1, training for 200 epochs, with the learning rate decaying by a factor of 0.1 at epochs 60, 120, and 180. For subsequent tasks, the base learning rate is set to 0.05, training for 100 epochs, with the learning rate decaying by a factor of 0.1 at epochs 45 and 90.
In the warm-start settings, we use the SGD optimizer with a batch size of 64. For the first task, the learning rate is set to 0.1, training for 100 epochs, with the learning rate decaying by a factor of 0.1 at epochs 45 and 90. For subsequent tasks, the base learning rate is set as follows: 0.01 for CIFAR100 (5 tasks) and TinyImageNet (5 tasks and 10 tasks), 0.03 for ImageNet100, and 0.05 for CIFAR100 (10 and 20 tasks) and TinyImageNet (20 tasks). Training lasts for 100 epochs on CIFAR100, with the learning rate decaying by a factor of 10 at epochs 45 and 90, and for 50 epochs on TinyImageNet and ImageNet100, with the learning rate decaying by a factor of 10 at epochs 35 and 45.
The weight decay is set to 5e-4 for the first task and 2e-4 for subsequent tasks.
During the incremental stages, we adapt the base learning rate based on the number of seen classes. The adaptive learning rate is defined as $lr_{adapt} = \frac{|C_t|}{|C_{1\colon t-1}|} \cdot lr$.
For noise parameter $\alpha$, after searching from [0.01, 0.05, 0.1, 0.2, 0.3, 0.4, 0.5], we select 0.2 as the optimal $\alpha$ value.

\section{More detailed pseudo-code}\label{sec:code}
We provide a more detailed pseudo-code for RoSE in \cref{alg:overall_detail}.
During training, we train the encoder using main loss $\mathcal{L}_{base}$ and the projector with auxiliary loss $\mathcal{L}_{aux}$.
During testing, we first initialize two queues, $Q_{old}$ and $Q_{new}$.
For each test, we update the projector by Eq. (12). 
The old prototypes are then fed into the updated projector to predict their new positions for image classification.
Leveraging information from recent test data, we can accurately estimate the drift of old semantics, ensuring that the prototypes remain aligned with the encoder.

\begin{algorithm}[!h]
    \caption{Overall Algorithm during task $t$}
    \label{alg:overall_detail}
    \renewcommand{\algorithmicrequire}{\textbf{Input:}}
    \renewcommand{\algorithmicensure}{\textbf{Output:}}
    \renewcommand{\algorithmiccomment}[1]{\hfill $\triangleright$ #1}
    \begin{algorithmic}[1]
        \Require train dataset $D_t$, test dataset $D_t^{test}$, encoder $f$ and its parameter $\theta_{t-1}$, projector $P$ and its parameter $W$, old prototypes $p_{C_{1:t-1}}^{t-1}$, queue budget S, total epoch $\mathcal{E}$, learning rate $\sigma$, noise coefficient $\alpha$. 
        \Ensure classification results $res$    
        
        \State  $\text{copy}: \theta_t \leftarrow \theta_{t-1}$
        \For{$e$ in $\mathcal{E}$} \Comment{Train start}
            \For{$x \in D_t$} \Comment{optimize $\theta_t$ and $W$}
                \State $ \theta_t \leftarrow \theta_t-\sigma*\nabla_{\theta} \mathcal{L}_{base}(x, \theta_t)$ 
                \State $ W \leftarrow W-\sigma * \nabla_{W} \mathcal{L}_{aux}(x, \theta_t, W)$ 
            \EndFor
        \EndFor
        \State $p_c^t \leftarrow \frac{1}{N_c}\Sigma_{n=1}^{N_c}f(x_n^c), \forall c \in C_t$ \Comment{Eq. (2)}
        \State $Q_{old}=[~], Q_{new}=[~], res=[~]$ \Comment{Test start}
        \While{$|Q_{old}| \textless S$} \Comment{Fig. 2A}
            \State $\text{randomly select old class}~c~ \text{from}~ C_{1:t-1}$
            \State $z_{old} \leftarrow p_c^{t-1} + \alpha * \mathcal{N}(0, 1), \text{add $z_{old}$ into $Q_{old}$}$
            \State $z_{new} \leftarrow P(z_{old}), \text{add $z_{new}$ into $Q_{new}$}$
        \EndWhile
        \For{$x_{test} \in D_t^{test}$} \Comment{Fig.2B and C}
            \State $z_{old} = f(x_{test}, \theta_{t-1}), z_{new}=f(x_{test}, \theta_t)$
            \State $\text{update}~Q_{old}~\text{by}~z_{old}, \text{update}~Q_{new}~\text{by}~z_{new}$ 
            \State $W \leftarrow (Q_{old}^\top\thinspace Q_{old})^{-1}Q_{old}^\top\thinspace Q_{new}$ \Comment{Eq. (12)}
            \State $p_c^t \leftarrow P(p_c^{t-1}), \forall c \in C_{1:t-1}$ \Comment{Eq. (9)}
            \State $c^* \leftarrow \mathop{argmin}\limits_c~\text{cos}~\langle z_{new}, p_c^t\rangle, \forall c \in C_{1:t-1}$
            \State $\text{add}~c^*~\text{into}~res$
        \EndFor
        \State \Return $res$
    \end{algorithmic}
\end{algorithm}

\begin{table*}[]
\centering
\begin{tabular}{lccccccc} \toprule
\multirow{2}{*}{Method} & \multicolumn{3}{c}{CIFAR100} & \multicolumn{3}{c}{TinyImageNet} & ImageNet100 \\ \cmidrule(lr){2-8}
 & 5 tasks & 10 tasks & 20 tasks & 5 tasks & 10 tasks & 20 tasks & 10 tasks \\ \midrule
RoSE (Original) & 63.10 & 53.48 & 44.10 & 47.07 & 39.93 & 31.57 & 63.18 \\
RoSE (Eliminate) & 63.09 & 53.27 & 43.68 & 46.88 & 39.67 & 31.36 & 62.52 \\ \bottomrule
\end{tabular}
\caption{Comparison with eliminating the data leakage concern. ``RoSE (Eliminate)'' denotes that we have eliminated the potential data leakage concern of RoSE. }
\label{tab:eliminate}
\end{table*}

\begin{table}[]
\centering
\setlength{\tabcolsep}{1pt}
\resizebox{\linewidth}{!}{
\begin{tabular}{m{2.5cm}<{}m{1cm}<{\centering}m{1.2cm}<{\centering}m{1.2cm}<{\centering}m{1cm}<{\centering}m{1.2cm}<{\centering}m{2cm}<{\centering}} \toprule
\multirow{2}{*}{Method} & \multicolumn{3}{c}{CIFAR100} & \multicolumn{2}{c}{TinyImageNet} & ImageNet100 \\ \cmidrule(lr){2-7}
 & 5 tasks & 10 tasks & 20 tasks & 5 tasks & 10 tasks & 10 tasks \\ \midrule
RoSE (Same) &63.40&53.80& 41.80 & 45.50 & 38.80 & 60.80 \\
RoSE (Distinct) &63.55&53.00& 41.20 & 45.30 & 38.50 & 61.00 \\ \bottomrule
\end{tabular}
}
\caption{Accuracy of RoSE when using same or distinct test data across different testing stages. }
\label{tab:eliminate2}
\end{table}
\section{Experiments that eliminate the data leakage concern}\label{sec:sm_notsame}

Given the limited size of the test set, we reuse the same test data for each class at each testing stage in our experiments.  This raises a potential concern about test set leakage, where test data from later tasks might have already been exposed in earlier testing stages.
Although the projector is discarded after each testing stage, the projected prototypes—--possibly retaining information from the test set—--are retained.
To eliminate this concern, we also discard the projected prototypes after each testing stage.
Implementing RoSE under this setting requires maintaining $t-1$ old models and the original prototypes for all seen classes during task $t$ testing. Additionally, $t-1$ projectors must be trained using the current test data to align feature spaces of tasks $1$ to $t-1$ with task $t$.
This introduces significant computational overhead. However, it's worth noting that this approach serves solely to confirm that data leakage is not a concern. In practical applications where test data varies across stages, such measures are unnecessary.
\cref{tab:eliminate}  compares the original RoSE results with those obtained after eliminating the concern of data leakage in the cold-start settings.
The results show that eliminating potential leakage has negligible impact, with declines nearly all under $0.5\%$.
This confirms that the experimental settings in the main text can adequately reflect RoSE's performance in real-world scenarios.

Another approach to simulate real-world scenarios is to distribute the test data across testing stages evenly, with distinct test samples for each task. 
We conduct experiments under the cold-start settings, comparing the results of using the same test samples across all stages versus distinct test samples, while keeping the number of test samples per stage constant.
We control that the test data used in the last task testing are the same for both conditions.
As shown in \cref{tab:eliminate2}, the performances under both conditions are comparable. This further demonstrates that RoSE's performance does not rely on data leakage.

\section{Comparison to settings relative to our method}\label{sec:relative_setting}

We note that some settings also explore methods for training the model at test time, but they are parallel research areas with key differences from ours.
Considering that there may be a gap between the testing domain and the training domain, Test-time training methods use test data to adapt the model to out-of-distribution (OOD) shifts by learning an auxiliary task during training and testing (different from our objective of reducing forgetting).
Based on this, Fully test-time adaptation does not impose any constraints on training, adapting models solely at test time.
Further, continual test-time adaptation recognizes that the distribution of test data may evolve. Training at evolving test data also leads to forgetting, but this forgetting is mainly manifested in the ability of adapting to old domains, rather than the class-incremental forgetting that we study.
These settings focus on domain adaptation, using OOD evaluation data, and the model training is not divided into multiple tasks.
In contrast, RoSE leverages test stages to mitigate class-incremental forgetting, so the test data remains in-distribution (ID) and the training is divided into multiple tasks for the model to learn sequentially (hence resulting in forgetting).
\cref{tab:settings} further highlights these distinctions, showcasing the novelty of RoSE more clearly.

\begin{table*}[]
\centering

\begin{tabular}{lcccccc} \toprule
Settings & Split train & Train update & Split test & Test update & Evaluation data & Examples \\ \midrule
Fine-tuning & \XSolidBrush & \Checkmark & \XSolidBrush & \XSolidBrush & ID & - \\
Test-time training & \XSolidBrush & \Checkmark & \XSolidBrush & \Checkmark & OOD & \cite{sun2020test, liu2021ttt++, gandelsman2022test} \\
Fully Test-time adaption & \XSolidBrush & \XSolidBrush & \XSolidBrush & \Checkmark & OOD & \cite{wang2020tent, zhaodelta} \\
Continual test-time adaptation & \XSolidBrush & \XSolidBrush & \Checkmark & \Checkmark & OOD & \cite{wang2022continual, niu2022efficient} \\
Continual learning & \Checkmark & \Checkmark & \XSolidBrush & \XSolidBrush & ID 
 & \cite{li2017learning, yu2020semantic, zhu2021prototype} \\ \midrule
RoSE(Ours) & \Checkmark & \Checkmark & \XSolidBrush & \Checkmark & ID & - \\ \bottomrule
\end{tabular}
\caption{Introduction to settings relative to our method. `Split train/test' indicates that training or testing is divided into multiple tasks and learned sequentially. `Train/Test update' denotes the stages of interest for each setting.}
\label{tab:settings}
\end{table*}

\section{More experiments about the comparison with test-time updating methods}\label{sec:more_ttt_compare}
We provide the results of RoSE and other test-time updating methods across all cold-start and warm-start settings in \cref{tab:alltttb0b50}.
RoSE consistently outperforms them by a large margin, demonstrating its superior design for NECIL.

\begin{table*}[]
\centering
\begin{tabular}{lcccccccc}
\toprule
\multicolumn{8}{c}{\textbf{Cold-start}} \\ \midrule
 \multirow{2}{*}{Methods} & \multicolumn{3}{c}{CIFAR100} & \multicolumn{3}{c}{TinyImageNet} & ImageNet100 \\
 & \multicolumn{1}{c}{5 tasks} & \multicolumn{1}{c}{10 tasks} & \multicolumn{1}{c}{20 tasks} & \multicolumn{1}{c}{5 tasks} & \multicolumn{1}{c}{10 tasks} & \multicolumn{1}{c}{20 tasks} & \multicolumn{1}{c}{10 tasks} \\ \midrule
TENT~\cite{wang2020tent} & 36.12 & 13.12 & 6.00 & 12.02 & 8.22 & 4.74 & 31.52 \\
TTT~\cite{sun2020test} & 49.33 & 36.93 & 24.92 & 35.53 & 24.35 & 12.37 & 30.66 \\
TTT++~\cite{liu2021ttt++} & 50.99 & 40.16 & 31.89 & 42.58 & 31.54 & 20.35 & 32.32 \\
CoTTA~\cite{wang2022continual} & 40.83 & 28.36 & 18.10 & 30.81 & 9.65 & 6.10 & 14.90 \\ 
EATA~\cite{niu2022efficient} & 57.30 & 46.87 & 32.33 & 43.43 & 35.34 & 24.61 & 54.90 \\ \midrule
 RoSE (Ours) & \textbf{63.10} &\textbf{53.48}  & \textbf{44.10} & \textbf{47.07} & \textbf{39.93} & \textbf{31.57} & \textbf{63.18} \\ \midrule
\multicolumn{8}{c}{\textbf{Warm-start}} \\ \midrule
\multirow{2}{*}{Methods} & \multicolumn{3}{c}{CIFAR100} & \multicolumn{3}{c}{TinyImageNet} & ImageNet100 \\
 & 5 tasks & 10 tasks & 20 tasks & 5 tasks & 10 tasks & 20 tasks & 10 tasks \\ \midrule
TENT~\cite{wang2020tent} & 11.58 & 6.07 & 4.53 & 7.02 & 7.38 & 2.91 & 7.62 \\
TTT~\cite{sun2020test} & 53.69 & 45.55 & 27.93 & 40.46 & 34.22 & 24.34 & 34.39 \\ 
TTT++~\cite{liu2021ttt++} & 56.81 & 54.72 & 38.96 & 46.82 & 43.32 & 35.37 & 61.80 \\
CoTTA~\cite{wang2022continual} & 54.57 & 43.92 & 27.10 & 50.10 & 40.93 & 32.18 & 53.40\\
EATA~\cite{niu2022efficient} & 55.88 & 49.65 & 33.36 & 30.28 & 34.05 & 24.68 & 54.42 \\ \midrule
RoSE (Ours) & \textbf{66.70} & \textbf{64.46} & \textbf{59.54} & \textbf{54.64} & \textbf{51.11} & \textbf{50.07} & \textbf{69.62} \\
\bottomrule
\end{tabular}
\caption{Full comparison with other methods on cold-start and warm-start settings.
\textbf{Bold} indicates the best result. \underline{Underline} indicates the second-best result.
}
\label{tab:alltttb0b50}
\end{table*}

\section{Comparison with the gradient descent upper bound.}\label{sec:upper}
\cref{tab:oracle} compares our analytical solution approach (Ours) with the upper bounds of the gradient descent (GD$_{oracle}$) under cold-start 10 tasks settings on three benchmark datasets: CIFAR100, TinyImageNet, and ImageNet100.
The GD$_{oracle}$ method assumes prior access to the complete test dataset for offline projector optimization through gradient descent until convergence.
However, this scenario is impractical in real-world applications as it violates the fundamental assumption of testing where test data arrives sequentially and cannot be preprocessed collectively.
Notably, RoSE demonstrates superior performance over GD$_{oracle}$ on CIFAR100 and TinyImageNet, indicating that the analytical solution can achieve performance that is challenging for GD, effectively circumvent the insufficient optimization in online training.

\begin{table}[]
\centering
\begin{tabular}{m{1.1cm}<{}m{1.4cm}<{\centering}m{1.8cm}<{\centering}m{1.8cm}<{\centering}} \toprule
 & CIFAR100 & TinyImageNet & ImageNet100 \\ \midrule
Ours & 53.48 & 39.93 & 63.18 \\
GD$_{oracle}$ & 53.43 & 39.81 & 63.32 \\
\bottomrule
\end{tabular}
\caption{Comparison with gradient descent and upper bounds.}
\label{tab:oracle}
\end{table}

\begin{table*}[!h]
\centering
\begin{tabular}{ccccccc} \toprule
\multicolumn{7}{c}{\textbf{Cold-start}} \\ \midrule
\multicolumn{3}{c}{CIFAR100} & \multicolumn{3}{c}{TinyImageNet} & ImageNet100 \\ \midrule
5 tasks & 10 tasks & 20 tasks & 5 tasks & 10 tasks & 20 tasks & 10 tasks \\ \midrule
62.65$\pm$0.51 & 53.64$\pm$0.71 & 44.11$\pm$0.14 & 48.49$\pm$1.08 & 41.79$\pm$1.45 & 32.82$\pm$1.00 & 63.31$\pm$0.94 \\ \midrule
\multicolumn{7}{c}{\textbf{Warm-start}} \\ \midrule
\multicolumn{3}{c}{CIFAR100} & \multicolumn{3}{c}{TinyImageNet} & ImageNet100 \\ \midrule
5 tasks & 10 tasks & 20 tasks & 5 tasks & 10 tasks & 20 tasks & 10 tasks \\ \midrule
66.85$\pm$0.15 & 64.34$\pm$0.12 & 59.19$\pm$0.37 & 54.31$\pm$0.38 & 51.11$\pm$0.12 & 50.02$\pm$0.52 & 69.25$\pm$1.08 \\ \bottomrule
\end{tabular}
\caption{
Results with multiple runs.
}
\label{tab:statistic}
\end{table*}
\section{Experiments with multiple runs}\label{sec:sm_multiple_runs}
To validate the robustness of RoSE, we conduct three trials with different random seeds. The average results for both cold-start and warm-start scenarios are shown in \cref{tab:statistic}.


\end{document}